\documentclass{svproc}
\usepackage{cite}
\usepackage{amsmath,amssymb,amsfonts}
\usepackage{algorithmic}
\usepackage{graphicx}
\usepackage{textcomp}
\usepackage{xcolor}
\usepackage{subcaption}
\usepackage{url}
\usepackage[T1,OT1]{fontenc}

\begin{document}
\mainmatter              
\title{\LARGE \bf Custom PX4 firmware for autonomous hybrid aerial-marine missions\\}
\titlerunning{PX4 hybrid firmware}  
%
\author{Andrea Capuozzo, Fabio Ruggiero, Vincenzo Lippiello}
\authorrunning{A. Capuozzo et al.} 
%
%
\institute{PRISMA Lab, Department of Electrical Engineering and Information Technology, University of Naples Federico II, Via Claudio 21, Naples, 80125, Italy\\
Corresponding author's email: \email{andrea.capuozzo@unina.it}\\}

\maketitle              

\begin{abstract}
Mapping and monitoring aquatic environments can benefit from hybrid aerial–amphibious drones able to combine flight and water-surface navigation within the same mission. This paper presents a PX4 firmware extension for such platforms, introducing manual and autonomous marine navigation modes integrated with the standard PX4 mission pipeline and QGroundControl interface. The proposed framework preserves existing flight functionalities and safety mechanisms while enabling unified planning and execution of hybrid aerial–marine missions with differentiated aerial and marine waypoints. Simulated case studies validate the implementation and demonstrate stable surface navigation under calm and wavy conditions.
\end{abstract}
\section{INTRODUCTION}
Mapping and monitoring aquatic environments are costly operations that require skilled personnel, specialized equipment, dedicated support infrastructures, and significant logistical resources. In addition, traditional inspection campaigns expose operators to safety risks and can negatively impact the investigated ecosystems, for instance, through water and noise pollution caused by support vessels. Advances in robotics~\cite{robots_for_environment,aerial_monitoring} helped mitigate these limitations by allowing operations to be conducted in safer, more efficient, and less invasive ways.

Hybrid aerial--amphibious drones, also referred to as unmanned aerial--aquatic vehicles (UAAVs), represent a versatile class of robotic platforms operating both in flight and on the water surface within a unified design. UAAVs can combine aerial inspection of target areas with water-based operations, including landing to deploy probes, collect samples, perform close-range measurements, and navigate on the surface. Despite the increasing interest in hybrid robotic platforms, open-source autopilot software—such as PX4~\cite{PX4}, ArduPilot~\cite{Ardupilot}, iNav~\cite{iNav}, or Betaflight~\cite{betaflight}—primarily targets single-domain vehicles and supports aerial and underwater systems as separate configurations. As a result, their capability to integrate aerial flight and water-surface navigation within a unified control and mission-execution architecture remains limited.

This paper addresses this limitation by introducing custom modules and control modes within the PX4 firmware, enabling full integration with its existing functionalities while preserving the standard control architecture. The PX4 firmware\footnote{The custom version presented in this work is based on the release v1.15.0} was selected due to its modular structure, compatibility with multiple flight-controller boards, and permissive Berkeley Software Distribution (BSD) license, which allows firmware extensions without mandatory upstream contribution.
The proposed code is freely available to the community\footnote{\url{https://github.com/prisma-lab/PX4-HybridAerialMarineCustom}}

The remainder of this paper is organized as follows. Section~\ref{sec:sota} discusses related work on marine survey drones and autopilot firmware frameworks. Section~\ref{sec:fa} presents the firmware customization and navigation mode design. Section~\ref{sec:method} describes the control strategies for hybrid aerial--marine operations. Section~\ref{sec:sim} reports results from simulated scenarios, while Section~\ref{sec:disc} discusses the findings. Finally, Section~\ref{sec:end} concludes the paper and outlines future work.

\section{RELATED WORKS\label{sec:sota}}
Several approaches have been proposed in the literature to employ aerial and surface robotic platforms in marine inspections. Unmanned aerial vehicles (UAVs) are widely adopted as rapid and cost-effective tools for large-scale inspection and mapping of marine habitats~\cite{soa_1}, leveraging their ability to efficiently cover extensive areas from an elevated viewpoint. Complementarily, unmanned surface vehicles (USVs) and unmanned underwater vehicles (UUVs) are used to perform close-range inspection and data collection tasks. For instance, a cooperative system composed of a UAV, a UUV, and a USV to inspect floating structures from multiple perspectives before allowing manned vessels is presented in~\cite{soa_2}.

Open-source USV platforms based on frameworks such as ArduPilot have demonstrated improvements in efficiency and safety for surface water sampling tasks~\cite{soa_3,soa_4}. These systems highlight the potential of robotic platforms to reduce operational costs while maintaining high-quality environmental data acquisition~\cite{soa_4}. More generally, the combination of aerial and surface viewpoints has been recognized as beneficial for enhancing situational awareness and navigation performance during marine inspection missions~\cite{soa_6,soa_7}.

Despite these advances, most existing solutions rely either on multi-vehicle cooperation, where distinct platforms operate under separate control architectures, or on hybrid platforms in which different dedicated control systems still handle the aerial and marine functionalities. For instance, in the multicopter--hovercraft system presented in~\cite{soa_last}, the two locomotion modes are managed through separate PX4-based flight-control architectures. A first example of a more integrated solution is represented by SailMAV~\cite{10923704}, a fixed-wing aerial--aquatic vehicle capable of alternating between flight and wind-driven sailing while relying on a single PX4-based onboard controller. However, its architecture is specifically developed around a fixed-wing flying--sailing configuration and the associated sailing dynamics and control requirements.

Overall, although the relevance of hybrid aerial--marine operations is increasingly recognized, firmware-level integration that unifies flight and marine navigation within a single control architecture remains largely unexplored, particularly for multirotor aerial--aquatic vehicles employing active propulsion for water-surface navigation. This motivates the development presented in this work.

\subsection{Contributions \label{subsec:cont}}
To the best of the authors' knowledge, a unified PX4 firmware extension for both aerial flight and marine surface navigation has not yet been addressed in the literature. This motivates the proposed firmware-level framework, which integrates aerial flight, water landing, marine navigation, takeoff, QGroundControl~\cite{QGC} mission planning, waypoint classification, and safety-mode switching within a single PX4 architecture.

The framework targets multirotor aerial--aquatic vehicles with independent marine propulsion. Its mission-management architecture is platform-independent, whereas the marine control is tailored to a two-fixed-thruster configuration, as described in Section~\ref{sec:uaav_spec}. The main contributions are: $(i)$ integration of the marine navigation module within the PX4 framework; $(ii)$ extension of the PX4 Navigator for hybrid aerial--marine mission execution; $(iii)$ PX4 implementation of manual and autonomous water-surface control strategies; and $(iv)$ integration of the proposed modes into QGroundControl for hybrid mission planning and aerial--marine waypoint differentiation.

\section{FIRMWARE ARCHITECTURE\label{sec:fa}}
\begin{figure*}[t!]
    \centering
    \begin{subfigure}{0.45\textwidth}
        \centering
        \includegraphics[width=1.1\textwidth]{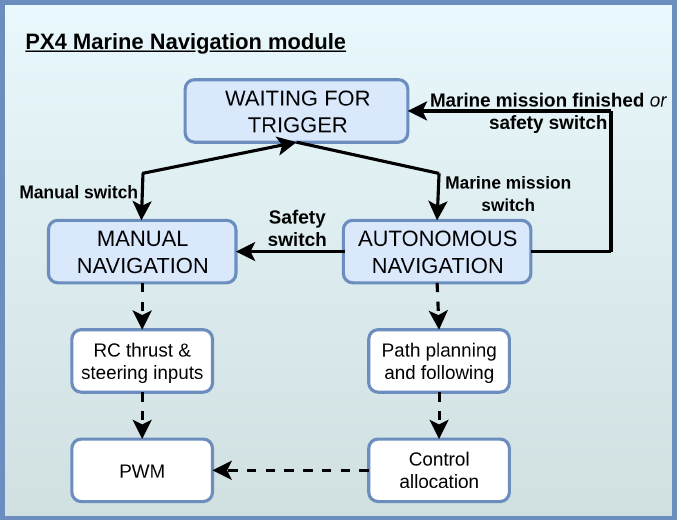}
        \caption{}
        \label{fig:fw_arch_a}
    \end{subfigure}
    \hfill
    \begin{subfigure}{0.45\textwidth}
        \hspace{-0.65cm}\includegraphics[width=1.1\textwidth]{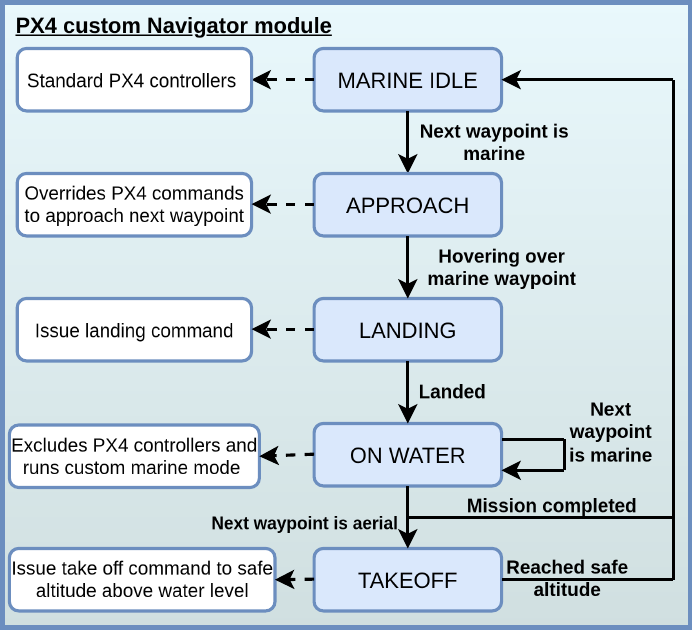}
        \caption{}
        \label{fig:fw_arch_b}
    \end{subfigure}
    \caption{State diagrams of: (a) the Marine Navigation module and (b) the customized Navigator module for marine mission branches. Blue boxes denote states, white boxes actions; solid and dashed lines indicate transitions and in-state actions, respectively.}
    \label{fig:fw_arch}
\end{figure*}

The proposed PX4 firmware extension follows the native modular architecture of PX4\footnote{\url{https://docs.px4.io/main/en/concept/architecture}} and introduces a dedicated \textit{Marine Navigation} module for surface operation. The module provides a unified framework for both manual and autonomous marine navigation, handling surface-specific guidance logic, mode transitions, and thruster actuation.

Moreover, the PX4 \textit{Navigator} module is extended to support hybrid aerial--marine missions, allowing smooth transitions between flight and marine--surface navigation phases. Dedicated interface updates expose the new modes in QGC, enabling intuitive mode selection and a clear distinction between aerial and marine waypoints.

\subsection{Marine Navigation module\label{subsec:mnm}}
The Marine Navigation module is the key component of the extended PX4 firmware and acts as a manager between the manual and autonomous marine navigation modes. It is implemented as a class inheriting from the following standard PX4 base classes:
\begin{itemize}
    \item \textit{ModuleBase}, which provides the core module functionalities;
    \item \textit{ModuleParams}, which enables the definition and access of control parameters;
    \item \textit{ScheduledWorkItem}, which schedules the execution of the module main function \texttt{run($\cdot$)} at a frequency of $50$~Hz.
\end{itemize}
All the information required for the module operation is exchanged through uORB\footnote{https://docs.px4.io/main/en/middleware/uorb} topics.

The state diagram of the module is shown in Fig.~\ref{fig:fw_arch_a}. When the system is started, the Marine Navigation module remains in a waiting state until a \texttt{vehicle\_control\_mode} uORB message signals the need for a navigation mode switch. The transition can be manually triggered by the pilot to enter or exit the marine manual navigation mode, or automatically requested by the Navigator module when a marine segment is required during mission execution, activating the autonomous marine navigation mode. During autonomous navigation, the pilot can always retake control—either aerial or marine manual navigation—through a safety switch. Otherwise, autonomous navigation continues until the marine path is completed.

In this work, the manual navigation mode is primarily intended as a safety mechanism that allows the operator to retake control of the UAAV during autonomous missions and is therefore designed for limited use.

In manual navigation mode, the custom module reads the inputs from the radio controller (RC) and directly converts them into pulse-width modulation (PWM) signals for thruster control. In autonomous navigation mode, path generation and path-following algorithms are executed to compute the thrust and torque commands to be applied to the UAAV, which are then transformed into PWM signals for thruster actuation through the appropriate control allocation matrix. The adopted control strategies for both modes are detailed in Section~\ref{sec:method}.

\subsection{Navigator module\label{subsec:nm}}
The PX4 Navigator module is responsible for autonomous flight modes and mission execution. The proposed customization enables the integration of marine segments within the mission execution pipeline. The main modifications were applied to the following files: \texttt{mission\_base.h}, \texttt{mission\_base.cpp}, \texttt{mission.h}, \texttt{mission.cpp}, and \texttt{mission\_block.cpp}. These changes implement the behavior described by the state diagram managing marine paths in Fig.~\ref{fig:fw_arch_b}.

Mission waypoints, selected through QGC, are classified as either aerial or marine based on a user-defined altitude threshold; waypoints below this altitude are considered marine. When an aerial waypoint is processed by the navigator module, the marine navigation logic remains idle and the standard PX4 controllers manage flight execution. Conversely, when a marine waypoint is received, the navigation logic transitions to an approach phase and temporarily overrides the PX4 standard controllers to guide the vehicle toward the destination waypoint while maintaining the previous altitude. Once the waypoint is reached, a landing command is issued and the UAAV performs a water landing. During this phase, the logic ensures that the vehicle is oriented toward the direction of the subsequent waypoint in order to avoid unnecessary maneuvers on the water surface. If the following waypoint is also classified as marine, the navigation mode switches to the autonomous marine mode described in Section~\ref{subsec:mnm} and remains active until a new aerial waypoint is received or the pilot retakes manual control, returning the state to idle (this transition may occur from any state). If the next waypoint is aerial, a takeoff command is issued and the UAAV climbs to a user-defined safe altitude above sea level before proceeding toward the waypoint, after which the logic returns to the marine idle state. The transition to the idle state also occurs when the mission is completed and no additional aerial or marine waypoints remain.

\subsection{QGroundControl interface}
The QGroundControl customizations introduced in this work\footnote{The custom version presented in this work is based on the release v4.3.0}, made publicly available to the community\footnote{{\fontencoding{T1}\selectfont
\url{https://github.com/prisma-lab/qgroundcontrol/tree/Stable_V4.3hybridUAV}}}, ensure that these core functionalities, described in Section~\ref{subsec:cont} are preserved. 

\begin{figure*}[t!]
    \centering
    \begin{subfigure}{0.35\textwidth}
        \centering
        \includegraphics[width=0.65\textwidth]{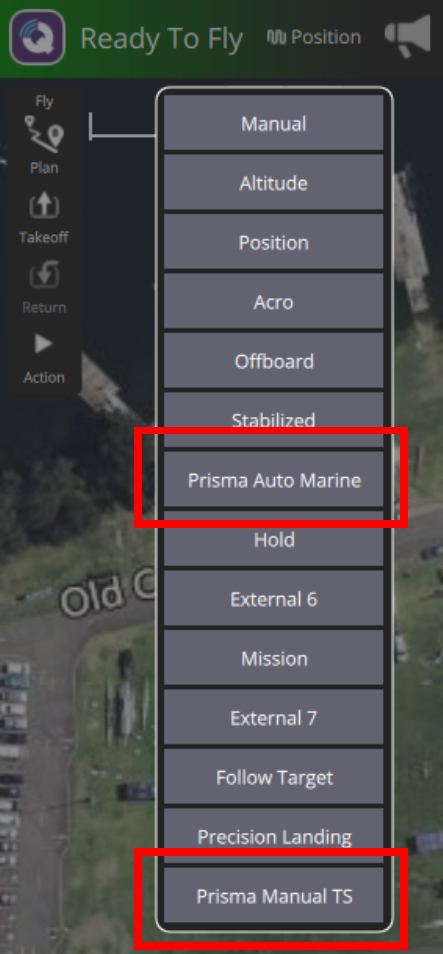}
        \caption{}
        \label{fig:QGC_int1}
    \end{subfigure}
    \hfill
    \begin{subfigure}{0.55\textwidth}
    \centering
        \includegraphics[width=0.9\textwidth]{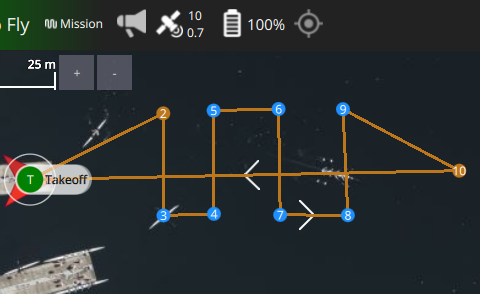}
        \caption{}
        \label{fig:QGC_int2}
    \end{subfigure}
    \caption{(a) QGC navigation mode selection showing the marine modes \textit{PRISMA Auto Marine} and \textit{PRISMA Manual TS}. (b) QGC mission plan with aerial and marine (in blue) waypoints.}
    \label{fig:QGC_interface}
\end{figure*}

The first customization enables the selection of the newly introduced flight modes directly from the navigation mode menu, as shown in Fig.~\ref{fig:QGC_int1}: \textit{PRISMA Auto Marine} for autonomous marine navigation and \textit{PRISMA Manual TS} for manual marine navigation. This functionality is implemented on the PX4 side rather than within QGC itself, by assigning valid \texttt{nav\_state} identifiers\footnote{https://docs.px4.io/main/en/msg\_docs/VehicleStatus} to the navigation modes and ensuring that the corresponding mode names are correctly communicated to QGroundControl\footnote{Handled through the file \texttt{PX4-Autopilot/src/lib/modes/ui.hpp}}.

The second customization allows users to quickly distinguish between aerial and marine waypoints. When QGroundControl is launched and a mission is uploaded, waypoint altitudes are evaluated and, if a value falls below the threshold defined in Section~\ref{subsec:nm}, the waypoint color is changed to blue (see Fig.~\ref{fig:QGC_int2}).

\section{METHODOLOGY\label{sec:method}}
Before discussing the mathematical details of how the marine navigation control modes are implemented, it is necessary to state some preliminary assumptions.

\subsection{UAAV specification\label{sec:uaav_spec}}
\begin{figure*}[t!]
    \centering
    \begin{subfigure}{0.45\columnwidth}
        \centering
        \includegraphics[width=\textwidth]{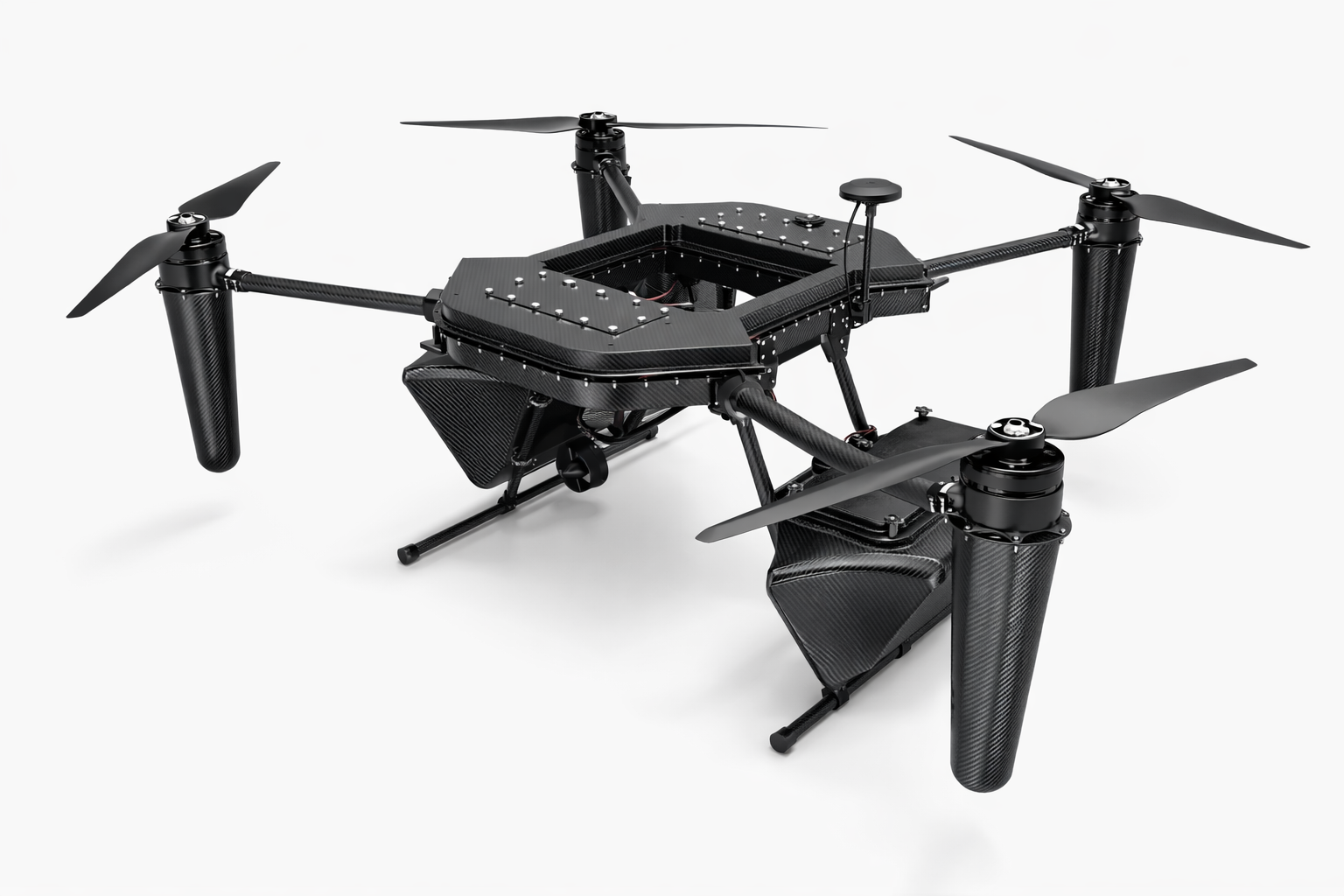}
        \caption{}
        \label{fig:drone_render}
    \end{subfigure}
    \hfill
    \begin{subfigure}{0.45\columnwidth}
    \centering
        \hspace{-0.7cm}\includegraphics[width=\textwidth]{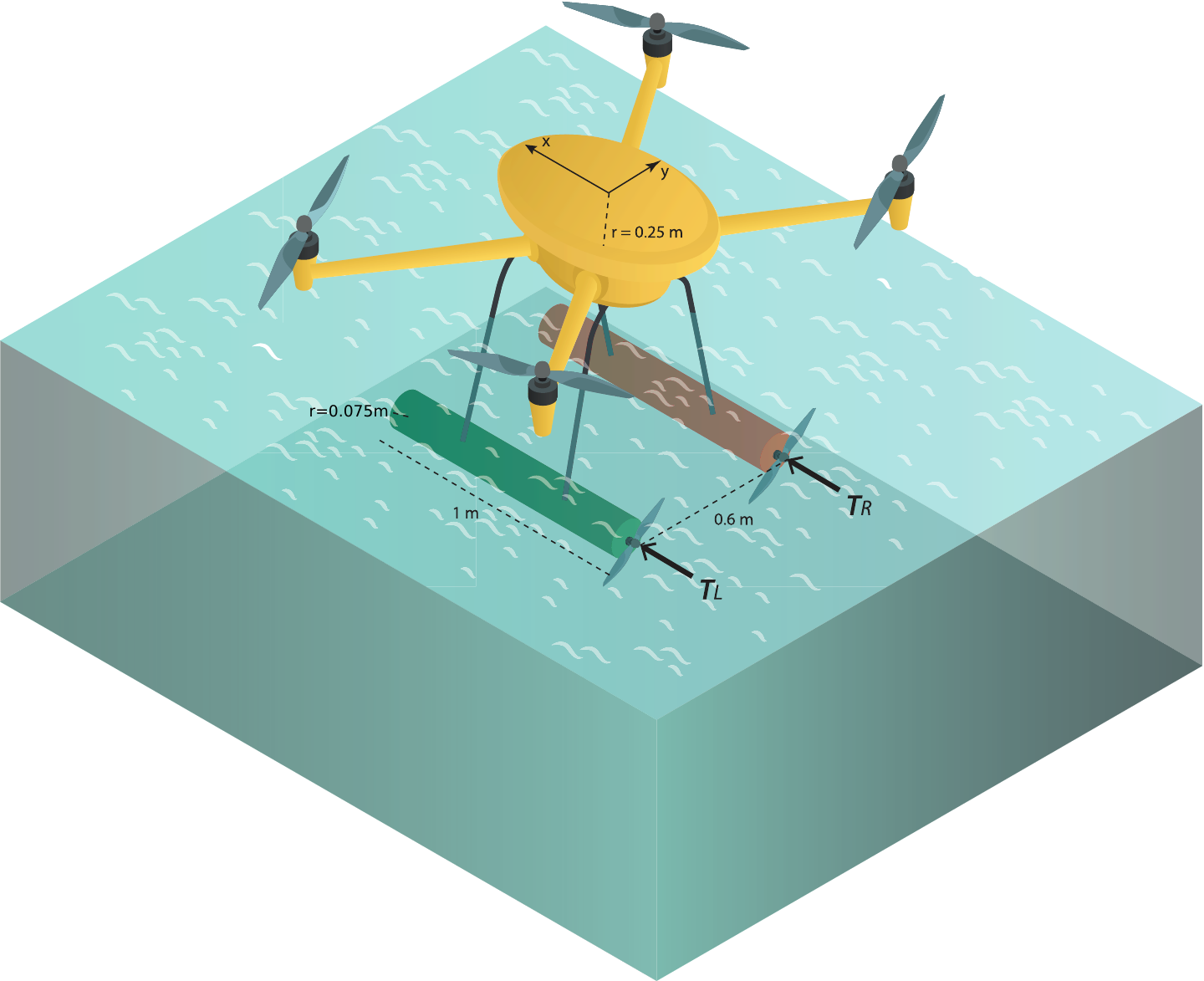}
        \caption{}
        \label{fig:drone_sim}
    \end{subfigure}
    \caption{(a) Rendering of the UAAV reference model used in this work. (b) Simplified UAAV model, used in simulations, showing the body reference frame, the thrust generated by the marine propellers, and the dimensions of the main components.}
    \label{fig:droneANDSim}
\end{figure*}

The UAAV reference platform considered in this work is illustrated in Fig.~\ref{fig:drone_render}. The vehicle consists of a quadrotor equipped with two buoyant hulls and two fixed, non-steerable marine propellers for surface navigation. A dedicated payload bay is located on the upper structure, allowing the integration of mission-specific equipment and sensing tools.

\subsection{Control allocation for marine vessels}
In low-speed applications, a three-degrees-of-freedom model is sufficient to describe the UAAV motion, accounting for surge (motion along the $x$-axis of the body frame), sway (motion along the $y$-axis of the body frame), and yaw (rotation about the $z$-axis of the body frame denoted by the angle  $\psi \in [-\pi\mbox{, }\pi)$), leading to the generalized force vector $\tau = [f_x,f_y,n]^T \in \mathbb{R}^3$, where $f_x, f_y \in \mathbb{R}$ denote the forces applied along the $x-$ and $y-$axis of the body frame, respectively, and $n \in \mathbb{R}$ denotes the torque about the $z$-axis of the body frame. In~\cite{allocation}, it is shown that, in the general case of a vessel equipped with $k>0$ marine propellers—either steerable, allowing the thrust direction to be varied, or fixed—each generating a thrust $T_i \in \mathbb{R}$ with $i=1, \ldots, k$, with positive and negative values denoting forward and backward thrust, respectively, and located at position $p_i = [\ell_{i,x},\ell_{i,y}]^T \in \mathbb{R}^2$, the generalized forces can be expressed as
\begin{equation} \label{eq:general_allocation}
    \tau = \begin{bmatrix}
        \cos\alpha_1 & \dots & \cos \alpha_k \\
        \sin \alpha_1 & \dots & \sin \alpha_k \\
        \ell_1 & \dots & \ell_k
    \end{bmatrix} T \mbox{,}
\end{equation}
where $\alpha_i \in [-\pi\mbox{, }\pi)$ is the orientation of the propeller $p_i$, $T = [T_1,\dots, T_k]^T \in \mathbb{R}^k$, and $\ell_i = \ell_{i,x}\sin\alpha_i - \ell_{i,y}\cos\alpha_i$.

As shown in Fig.~\ref{fig:droneANDSim}, where the configuration of the reference UAAV platform is illustrated, the vessel layout includes only two non-steerable marine propellers. This simplifies~\eqref{eq:general_allocation} into
\begin{equation} \label{eq:simp_allocation}
    \bar\tau = \begin{bmatrix}
        1 & 1 \\
        -\ell_{1,y} & -\ell_{2,y}
    \end{bmatrix} \bar{T} \mbox{,}
\end{equation}
with $\bar\tau = [f_x \mbox{, } n]^T \in \mathbb{R}^2$ and $\bar{T} = [T_\mathrm{L}\mbox{, } T_\mathrm{R}]^T$, as the thrust generated by the left and right propellers, respectively.

\subsection{Manual navigation control}
The manual navigation control is implemented as a feedforward strategy modulating each propeller's thrust and providing the pilot with the perception of independent control over total thrust and a virtual rudder. This design enables precise maneuvering, including pivoting without requiring forward motion.

The controller reads the radio control (RC) inputs through the uORB topic \texttt{manual\_control\_setpoint}, specifically the \texttt{throttle} and \texttt{roll} fields of the received message, interpreting them as total thrust $t$ and steering $s$ values, respectively, with $t \mbox{, } s \in [-1,1]$. In particular, $s \in [-1,0]$ corresponds to left steering, whereas $s \in [0,1]$ corresponds to right steering. The thrust commands to be generated by each propeller are then computed as
\begin{equation}
\begin{aligned}
    &T_\mathrm{L} = \begin{cases}
        t & \mbox{if } s \geq 0 \\
        t(1+2s) & \mbox{if } s < 0
    \end{cases} \mbox{ ,}
    \\ 
    &T_\mathrm{R} = \begin{cases}
        t(1-2s) & \mbox{if } s > 0 \\
        t & \mbox{if } s \leq 0
    \end{cases} \mbox{ .}
\end{aligned}
\end{equation}
Note that $T_\mathrm{L} \mbox{, } T_\mathrm{R} \in [-1 \mbox{, } 1]$ since they represent normalized thrust values. 

\subsection{Autonomous navigation control}
The autonomous navigation controller adopts a path-following strategy for underactuated marine vehicles based on the hand-position framework and input–output feedback linearization presented in~\cite{path_following}. The controlled output is a virtual point $h=[\xi_1\mbox{, }\xi_2]^T \in \mathbb{R}^2$, referred to as the hand-position point, located at a fixed distance $l$ ahead of the UAAV center of mass $p=[x\mbox{, }y]^T \in \mathbb{R}^2$ along the $x$ axis of the body frame
$\xi_1 = x + l\cos\psi$ and $\xi_2 = y + l\sin\psi$.
The choice of controlling the hand-position point introduces an additional controllable output: a surge actuation applied at $p$ generates a surge motion of $h$, while a yaw actuation induces a lateral motion of $h$ proportional to $l$. This property enables the application of input–output feedback linearization to underactuated marine platforms.

Applying the coordinate transformation and input mapping described in~\cite{path_following}, and denoting by $\xi_3,\xi_4 \in \mathbb{R}$ the relative velocity components of $h$, by $V_x, V_y \in \mathbb{R}$ the unknown ocean current components expressed in the world frame, and by $u_1, u_2 \in \mathbb{R}$ the virtual control inputs, the external dynamics can be written as
\begin{equation}
\begin{aligned}
    \dot\xi_1 &= \xi_3 + V_x,\\
    \dot\xi_2 &= \xi_4 + V_y,\\
    \dot\xi_3 &= u_1,\\
    \dot\xi_4 &= u_2.
\end{aligned}
\end{equation}

The virtual inputs are related to the generalized force vector $\bar{\tau}$ through the feedback-linearization mapping
\begin{equation}\label{eq:from_u_to_tau}
\bar{\tau} =
\begin{bmatrix}
    \cos\psi & -l\sin\psi \\
    \sin\psi & \;\;l\cos\psi
\end{bmatrix}^{-1}
\begin{bmatrix}
    u_1 - F_{\xi,1}(\cdot) \\
    u_2 - F_{\xi,2}(\cdot)
\end{bmatrix},
\end{equation}
with $F_{\xi,1}(\cdot),F_{\xi,2}(\cdot) \in \mathbb{R}$ the nonlinear drift dynamics of the hand-point acceleration.

The virtual inputs $u_1$ and $u_2$ are designed for a generic $C^2$ path $\gamma(s)=(x(s),y(s))$, where $s\geq 0$ denotes the arc-length parameter, and include feedforward terms accounting for its local curvature~\cite{path_following}. In this work, the marine path is defined as a sequence of straight-line segments connecting consecutive waypoints, resulting in a path with zero curvature and constant tangent direction. Under these assumptions, curvature-dependent feedforward terms vanish, leading to the following simplified choice of the virtual inputs
\begin{equation}
\begin{aligned}
    u_1 &= -k_{vx}\big(\xi_3-\dot{x}(s)\big)
          -k_{px}\big(\xi_1-x(s)\big)
          -k_{Ix}I_x,\\
    u_2 &= -k_{vy}\big(\xi_4-\dot{y}(s)\big)
          -k_{py}\big(\xi_2-y(s)\big)
          -k_{Iy}I_y,
\end{aligned}
\end{equation}
where $\dot{x}(s)=\dot{s}t_x$ and $\dot{y}(s)=\dot{s}t_y$, with $[t_x\mbox{, }t_y]$ denoting the constant unit tangent of the active path segment. Following the path-following law in~\cite{path_following}, the actual path-progress rate $\dot{s}$ is obtained by modulating a nominal rate $\dot{s}_{\mathrm{nom}}$ according to the hand-point tracking error. The constants $k_{vx}$, $k_{vy}$, $k_{px}$, $k_{py}$, $k_{Ix}$, $k_{Iy} \in \mathbb{R}^+$ are the control gains. The integral terms are defined as
$I_x = \int_0^t \big(\xi_1(\tau)-x(s(\tau))\big)\,d\tau$ and  $I_y = \int_0^t \big(\xi_2(\tau)-y(s(\tau))\big)\,d\tau$, and provide steady-state rejection of constant environmental disturbances such as ocean currents.

Following~\cite{path_following}, the path-progress dynamics are regulated as a function of the hand-point tracking error, 
$e_h^2 = \big(\xi_1-x(s)\big)^2+\big(\xi_2-y(s)\big)^2$.
The path-progress dynamics are defined as
$\dot{s} = \dot{s}_{\mathrm{nom}}\left(1-\tanh(e_h^2)\right)$. The nominal rate $\dot{s}_{\mathrm{nom}}$ is reduced near the target waypoint by scaling it with $d_{\mathrm{wp}}/d_{\mathrm{slow}}$, where $d_{\mathrm{wp}}$ is the remaining waypoint distance and $d_{\mathrm{slow}}$ defines the slowdown region. Once $d_{\mathrm{wp}}<d_r$, with $d_r$ the waypoint acceptance radius, $\dot{s}_{\mathrm{nom}}$ is set to zero.
For small tracking errors, $\tanh(e_h^2)\approx 0$, $\gamma(s)$ advances at nearly the nominal rate. Conversely, when the error increases, $\tanh(e_h^2)\to 1$, the progression rate reduces. This $\gamma(s)$ error-dependent progression is particularly useful at waypoint transitions and during turning maneuvers, since it avoids the generation of an excessively distant moving target and allows the controller to re-establish alignment before proceeding along the next segment.

Once the virtual inputs are computed, the generalized forces $\bar{\tau}$ are obtained from~\eqref{eq:from_u_to_tau} and subsequently mapped to the individual propeller commands through the inversion of~\eqref{eq:simp_allocation}.

To account for steady ocean currents in the implementation, the hand-point
velocities are evaluated using water-relative kinematics. Specifically, the
measured ground velocity, estimated by the PX4 state estimator using GNSS and inertial measurements, is corrected by subtracting an online estimate of the
current velocity $\hat{v}_c \in \mathbb{R}^2$
\begin{equation}
\begin{bmatrix}
\dot x_r\\
\dot y_r
\end{bmatrix}
=
\begin{bmatrix}
\dot x\\
\dot y
\end{bmatrix}
-
\hat{v}_c .
\end{equation}
Accordingly, the relative velocity components of the hand-position point are computed as
\begin{equation}
\begin{aligned}
\xi_3 &= \dot x_r - l\sin\psi\,\dot\psi,\\
\xi_4 &= \dot y_r + l\cos\psi\,\dot\psi.
\end{aligned}
\end{equation}
The estimate $\hat{v}_c$ is updated online from the hand-point velocity
tracking error, which provides an indication of the persistent drift affecting
the vehicle motion. In compact form, this adaptation can be expressed as
\begin{equation}
\dot{\hat{v}}_c = k_c e_v - k_{cl}\hat{v}_c ,
\qquad
e_v =
\begin{bmatrix}
\xi_3-\dot{x}_d\\
\xi_4-\dot{y}_d
\end{bmatrix},
\end{equation}
where $k_c$ and $k_{cl}$ are positive adaptation and leakage gains, respectively.

In the proposed PX4 implementation, this current-compensation mechanism is
combined with an alignment-gated update strategy. The alignment metric is
computed as the cosine of the angle between the vehicle's ground velocity
direction and the tangent of the active path segment. Therefore, the current
estimate $\hat{v}_c$ and the integral action are updated only when this metric
exceeds the threshold $g_{\mathrm{align,th}}$ and the surge velocity is above a minimum threshold
$v_{th}$. This prevents unreliable compensation during turning maneuvers and
low-speed transients, where the tracking error is mainly caused by reorientation
and hydrodynamic effects rather than by persistent environmental disturbances.

\section{SIMULATIONS\label{sec:sim}}

\subsection{Simulation set up}
The simulations were carried out on a standard personal computer equipped with an \textit{Intel Core i7} ($11^{\mathrm{th}}$ generation) processor and $32$~GB of RAM, running \textit{Ubuntu 22.04}. The marine environment was simulated in the \textit{Virtual RobotX} (VRX) environment~\cite{c14} using \textit{Gazebo Garden v7.9.0}, while the UAAV dynamics and control architecture were executed through the \textit{PX4 Autopilot} firmware in software-in-the-loop (SITL) mode. In SITL mode, the PX4 flight stack executes the same firmware-level logic intended for onboard deployment, while Gazebo emulates the vehicle dynamics, sensors, and actuators. The simulations therefore provide a direct software-level validation of the proposed PX4 architecture and its aerial, marine, and transition-management functions. 

The buoyancy model was modified by representing the floating behavior as a mass--spring--damper system to include water damping effects during drone--sea interaction~\cite{c15}. This prevents persistent oscillations that would otherwise arise from a purely elastic buoyant restoring force. The simplified UAAV model used in the simulations is shown in Fig.~\ref{fig:drone_sim}. The UAAV parameters are: mass $m=20$ kg, buoyant-hull length $L=1$ m, distance between buoyant-hull centers $D=0.6$ m, buoyant-hull radius $R=0.075$ m, total height $h=0.375$ m, and maximum bidirectional marine-propeller thrust $T_{\max}=5$ N.
The hand-position distance, autonomous-navigation gains, nominal progression rate, and control thresholds were selected empirically through trial-and-error tuning to obtain stable path-following performance. The adopted values are
$l=1.7$~m, $k_{vx}=k_{vy}=1$, $k_{px}=k_{py}=0.7$, $k_{Ix}=k_{Iy}=0.05$, $\dot{s}_{\mathrm{nom}}=5$~m/s, $v_{\mathrm{th}}=1.2$~m/s, $d_{\mathrm{slow}}=3$~m, $d_r=0.5$~m, $k_c=0.7$, $k_{cl}=0.02$ and $g_{\mathrm{align,th}}=0.75$.

Two case studies are considered, both involving a hybrid aerial--marine mission in which the UAAV takes off from land, reaches a target area over water, lands, follows a boustrophedon path of eight QGC-selected marine waypoints, and then returns to the starting point. The two cases differ only in environmental conditions: calm sea in the first case and a Gerstner-wave field~\cite{c10} in the second.
The analysis focuses on the marine phase, assessing path-following performance and the ability of the UAAV to remain on track. The accompanying video\footnote{https://youtu.be/dyPZIZCPyYU} shows the full execution of the first mission, including pilot-initiated manual takeover for safety recovery.

\subsection{Case study 1)}

\begin{figure*}[t!]
    \centering
    \begin{subfigure}{0.45\textwidth}
        \centering
        \includegraphics[width=0.85\textwidth]{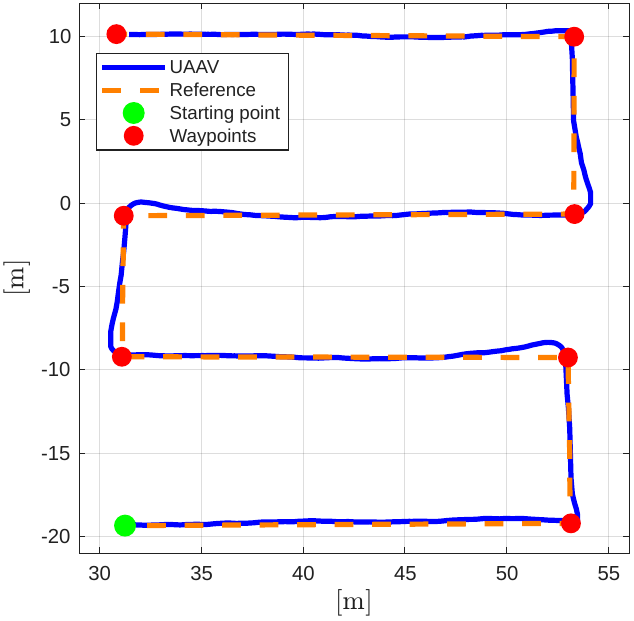}
        \caption{}
        \label{fig:calm_traj}
    \end{subfigure}
    \hfill
    \begin{subfigure}{0.45\textwidth}
    \centering
        \includegraphics[width=\textwidth]{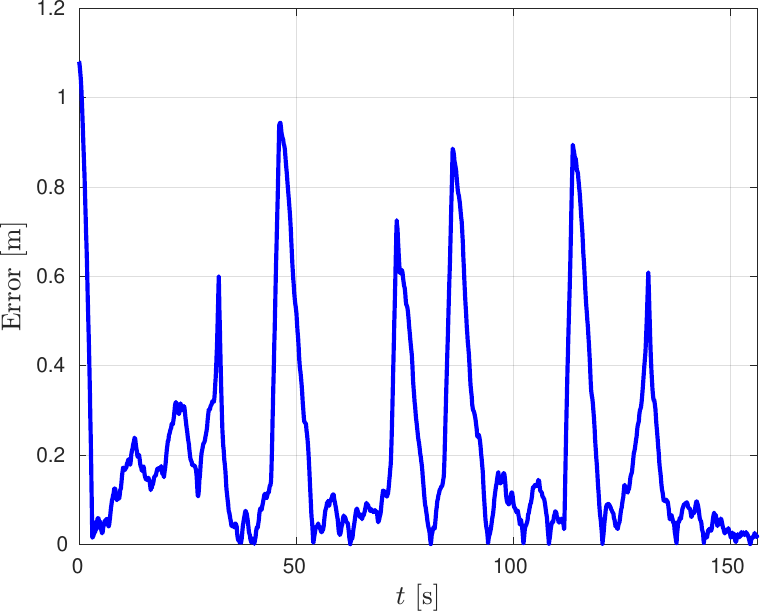}
        \caption{}
        \label{fig:calm_traj_error}
    \end{subfigure}
    \caption{\textit{Case study 1)}: (a) comparison between executed marine path and reference; (b) off-track error norm.}
    \label{fig:cs1}
\end{figure*}

\begin{figure*}[t!]
    \centering
    \begin{subfigure}{0.45\textwidth}
        \centering
        \includegraphics[width=\textwidth]{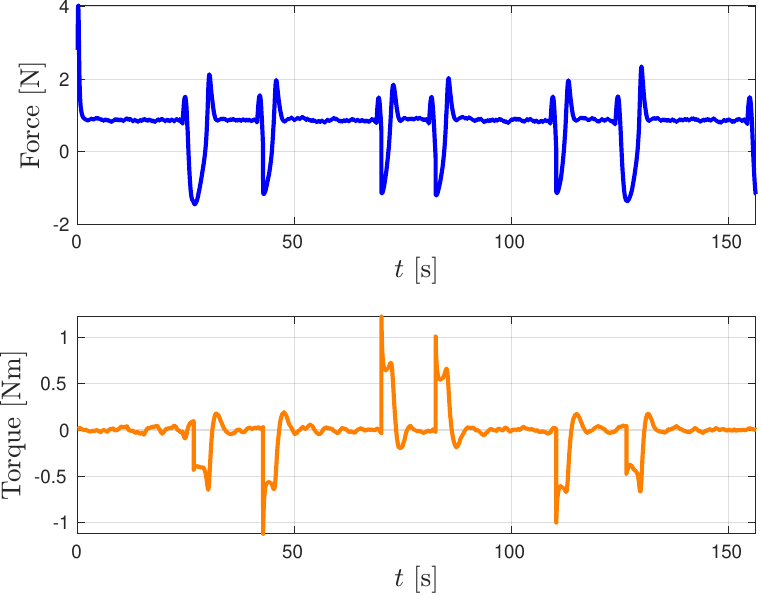}
        \caption{}
        \label{fig:c1f}
    \end{subfigure}
    \hfill
    \begin{subfigure}{0.45\textwidth}
    \centering
        \includegraphics[width=\textwidth]{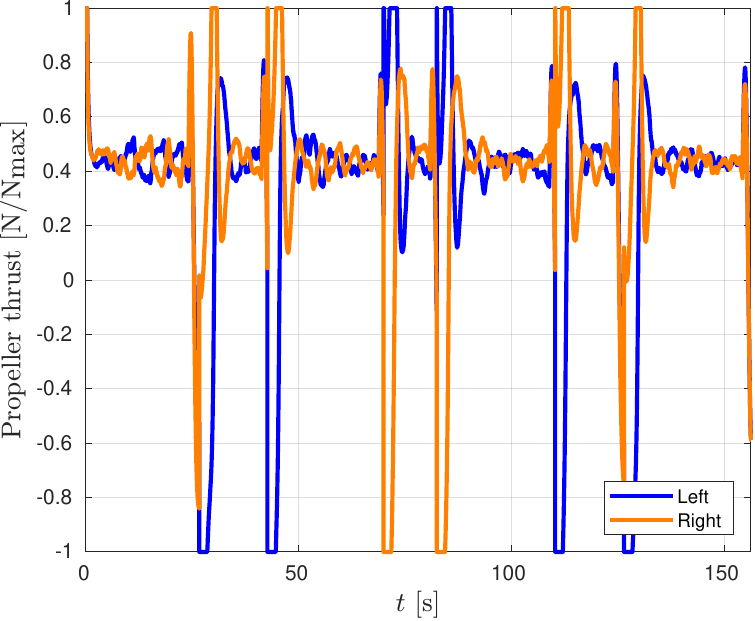}
        \caption{}
        \label{fig:c1t}
    \end{subfigure}
    \caption{\textit{Case study 1)}: (a) computed force and torque controls; (b) left and right normalized propeller control inputs.}
    \label{fig:calm_tau}
\end{figure*}

Figure~\ref{fig:calm_traj} reports the executed path and the off-track error norm, computed as the distance between the UAAV position and the reference position along the prescribed marine path. The vehicle follows eight waypoints arranged in a boustrophedon pattern. The error remains bounded, with peaks occurring mainly during turns, where inertial and hydrodynamic effects temporarily limit rapid reorientation. Its maximum value stays below the vehicle length, indicating that the deviation remains within one characteristic vehicle dimension and providing a normalized interpretation of tracking accuracy, consistent with ship-domain approaches based on length-dependent safety regions~\cite{shipdomain}. After each turn, the controller restores alignment with the reference path, yielding low errors along straight segments and localized peaks at segment transitions.

This behavior is further explained by Fig.~\ref{fig:calm_tau}, which shows the forces, torques, and propeller control inputs generated by the autonomous navigation controller. Near the end of each marine segment, forward thrust is reduced to ensure convergence to the waypoint; when motion toward the next waypoint starts, a stronger torque aligns the vessel with the new heading. This deceleration, reorientation, and re-acceleration sequence explains the transient turning errors and the subsequent recovery of path-following performance.

\subsection{Case study 2)}

\begin{figure*}[t!]
    \centering
    \begin{subfigure}{0.45\textwidth}
        \centering
        \includegraphics[width=0.85\textwidth]{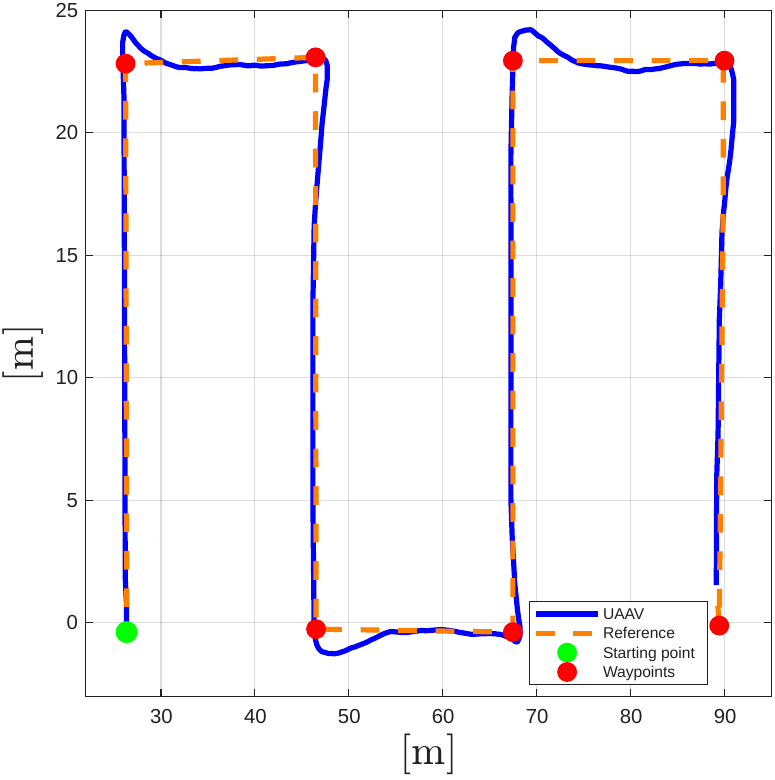}
        \caption{}
        \label{fig:c2traj}
    \end{subfigure}
    \hfill
    \begin{subfigure}{0.45\textwidth}
    \centering
        \includegraphics[width=\textwidth]{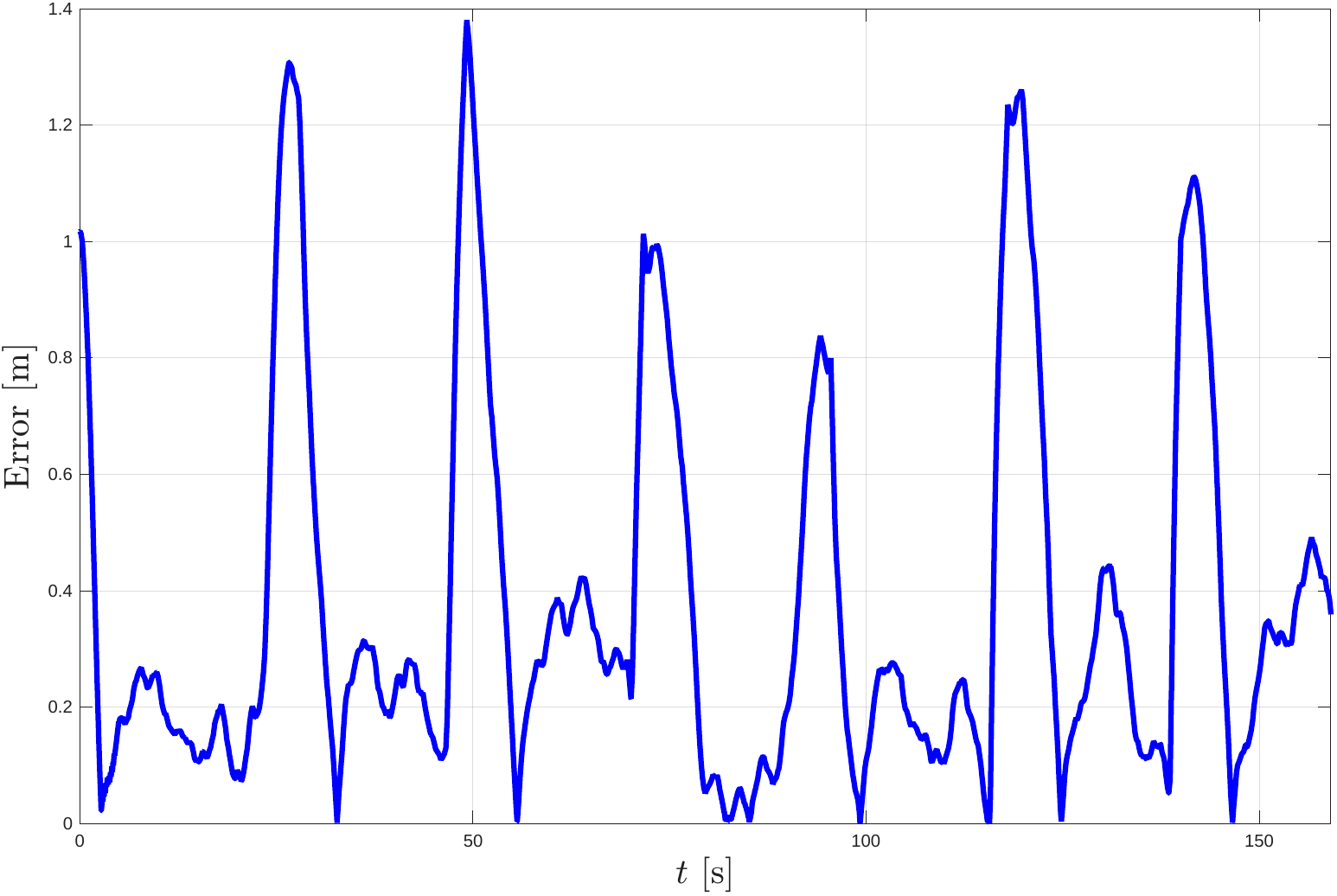}
        \caption{}
        \label{fig:c2err}
    \end{subfigure}
    \caption{\textit{Case study 2)}: (a) comparison between executed marine path and reference; (b) off-track error norm.}
    \label{fig:wave_traj}
\end{figure*}

\begin{figure*}[t!]
    \centering
    \begin{subfigure}{0.45\textwidth}
        \centering
        \includegraphics[width=\textwidth]{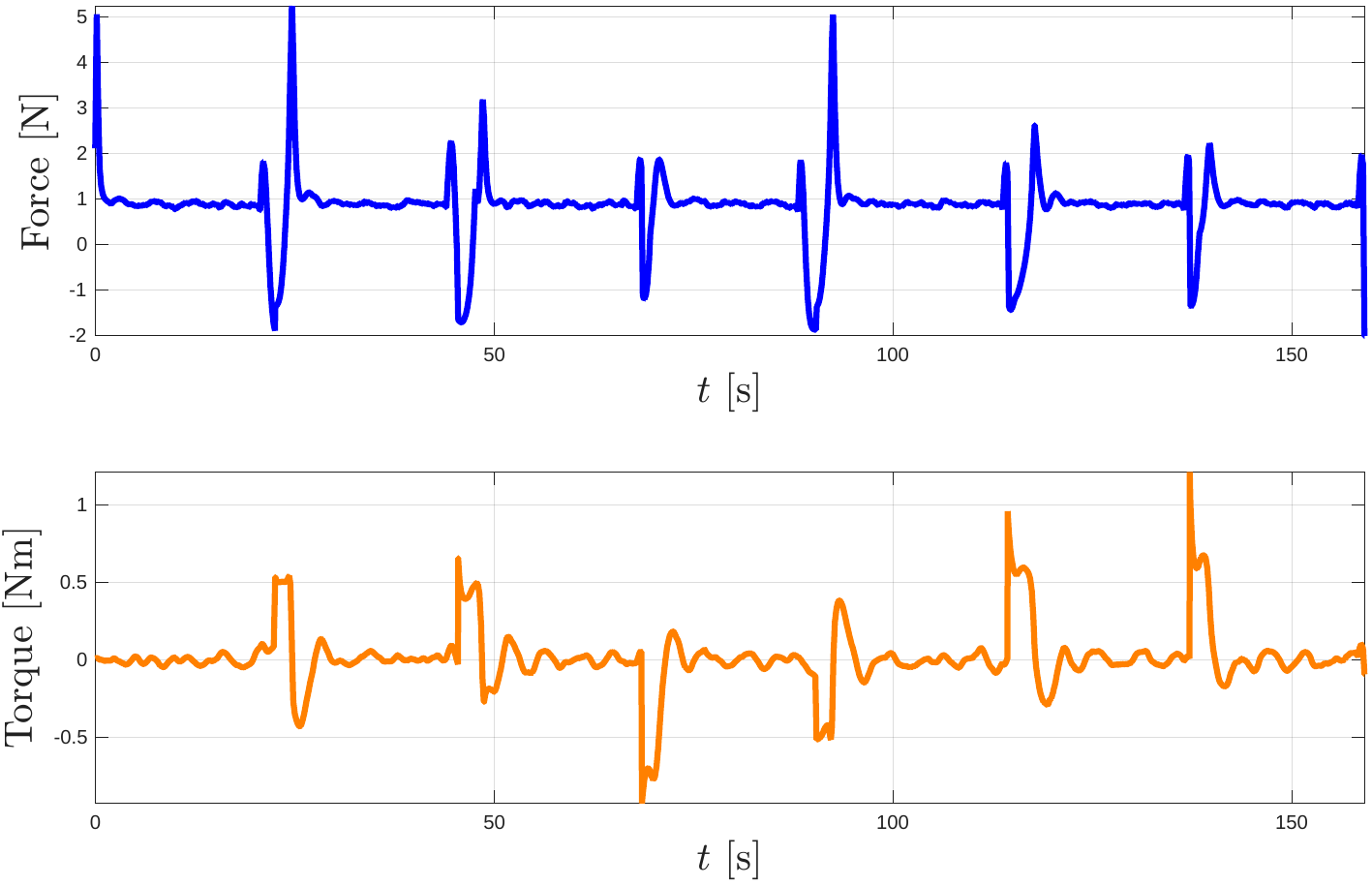}
        \caption{}
        \label{fig:c2f}
    \end{subfigure}
    \hfill
    \begin{subfigure}{0.45\textwidth}
    \centering
        \includegraphics[width=\textwidth]{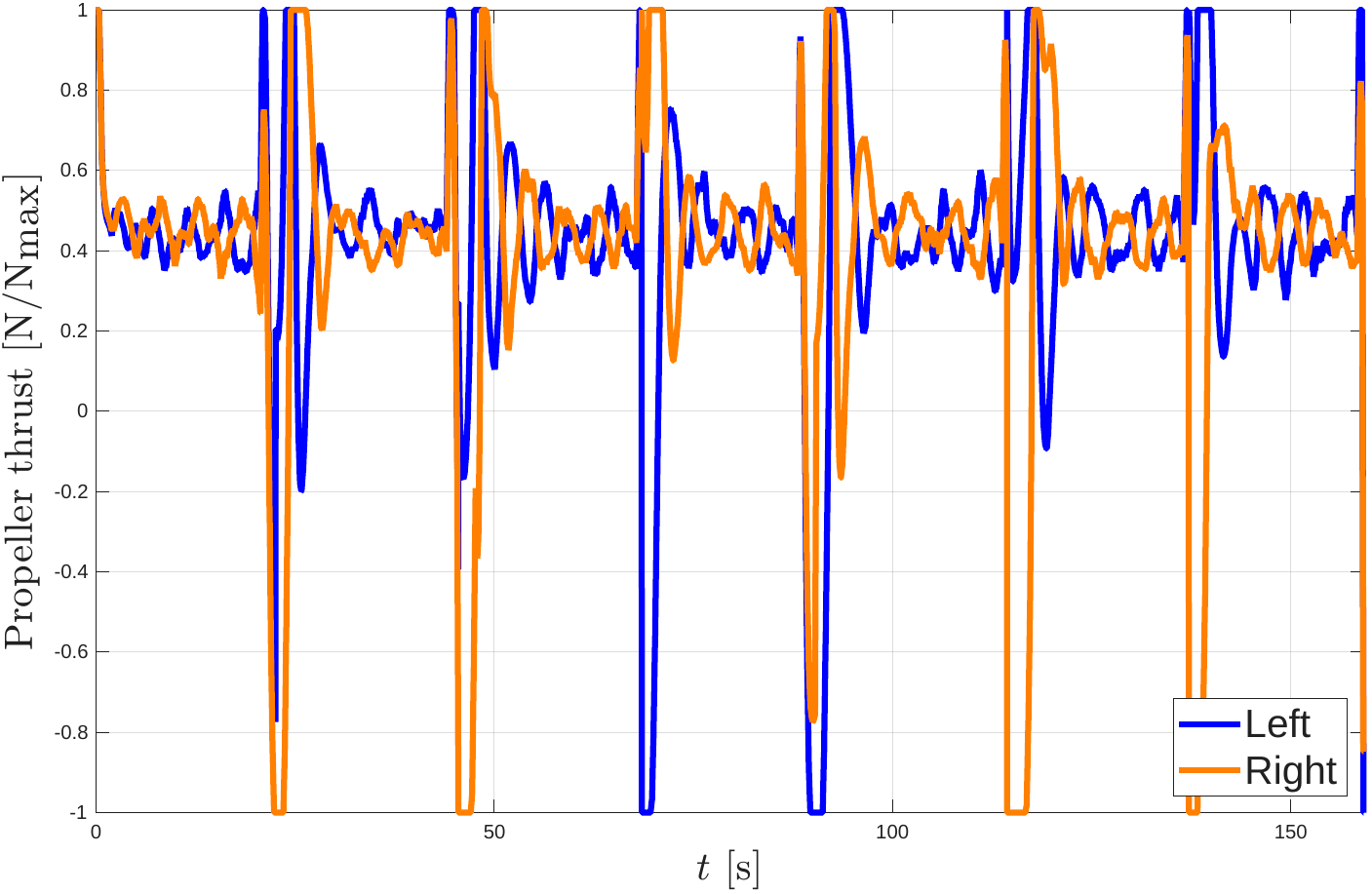}
        \caption{}
        \label{fig:c2t}
    \end{subfigure}
    \caption{\textit{Case study 2)}: (a) computed force and torque controls; (b) left and right normalized propeller control inputs.}
    \label{fig:wave_tau}
\end{figure*}

In this scenario, the UAAV again executes an eight-waypoint boustrophedon path. The wave field is composed of two incident wave components with a period of $4$~s, an angular separation of $0.4$~rad, and a median wave amplitude of $1.5$~m. Figure~\ref{fig:wave_traj} shows the executed path and the corresponding reference path. Overall, the tracking performance remains acceptable. Compared with \textit{case study 1)}, the error slightly increases along the straight segments, although it remains below the vehicle-length threshold in these portions of the path.

Larger deviations are observed during turning maneuvers, where the combined effect of wave-induced disturbances, vehicle inertia, and hydrodynamic transients makes pivoting less stable. In a few turns, the tracking error exceeds the vehicle-length threshold, indicating that the wave field mainly affects the most demanding phases of the path rather than the straight-line motion. Figure~\ref{fig:wave_tau} reports the resulting force and torque commands, together with the propeller control inputs. As expected, the wave disturbances produce more pronounced oscillations in the control signals compared with the calm-water case.

\section{DISCUSSION\label{sec:disc}}
The results in Section~\ref{sec:sim} highlight both the strengths and limitations of the proposed firmware extension and autonomous marine navigation control.

The modified firmware manages hybrid missions, including marine navigation phases and the required safety mechanisms. It allows rapid mode switching, enabling the operator to promptly regain control in emergency situations. However, hybrid missions are not yet checked for feasibility before execution, e.g., based on battery state, nor adapted online through actions such as automatic return-to-home when the remaining energy becomes insufficient. Moreover, the current framework supports only navigation-oriented marine tasks; future extensions should include richer surface-operation directives.

Regarding the control algorithm, simulations show stable navigation performance, also under non-ideal environmental conditions, while keeping the propeller inputs within the admissible normalized range. Further improvements could be achieved at the path-planning level, smoothing waypoint transitions and reducing the transient deviations during maneuvers.

Finally, the manual navigation mode could be improved, since its feedforward structure still requires skilled operators. Future work should introduce assistance mechanisms to reduce pilot workload and enable reliable water-surface control also by non-expert users.

\section{CONCLUSION AND FUTURE WORK\label{sec:end}}
This paper proposes a new PX4 firmware version for UAAVs operating in hybrid aerial--marine autonomous missions. The objective is to provide a unified control framework ensuring safe operation for the pilot, the vehicle, and the surrounding environment. The introduced marine control modes enable manual and autonomous navigation while preserving the standard user interface. Simulation results demonstrate the effectiveness of the extended firmware and its potential for further development. Future work will improve firmware capabilities and safety mechanisms through mission feasibility checks, online replanning, and additional specifications for hybrid mission execution. The marine control framework will also be extended through improved path planning and a more accessible manual control mode, together with hardware experiments to validate the proposed framework on the physical UAAV platform.

\section*{Acknowledgments}
The research leading to these results has been partially supported by \textit{Centro Nazionale 5} (CN5) - \textit{National Biodiversity Future Center} (NBFC)~- \textit{SPOKE~1} (Codice progetto MUR: CN0000033 CUP UNINA: E63C22000990007).
This research was also partially supported by the \textit{Space It Up} project funded by the \textit{Italian Space Agency} (ASI), and the \textit{Ministry of University and Research} (MUR) under contract n. 2024-5-E.0 - CUP n. I53D24000060005. 

The rendering of Fig.~\ref{fig:drone_render} has been obtained using AI tool such as OpenAI ChatGPT (v. 5.2) with simple prompt commands.

\bibliographystyle{IEEEtran}
\bibliography{biblio}

@article{robots_for_environment,
  author  = {Robert Bogue},
  title   = {The Role of Robots in Environmental Monitoring},
  journal = {Industrial Robot: The International Journal of Robotics Research and Application},
  year    = {2023},
  volume  = {50},
  number  = {3},
  pages   = {369--375},
  doi     = {10.1108/IR-12-2022-0316}
}

@incollection{aerial_monitoring,
  author    = {Daniele Ventura and Andrea Bonifazi and Maria Flavia Gravina and Gian Domenico Ardizzone},
  title     = {Unmanned Aerial Systems ({UAS}s) for Environmental Monitoring: A Review with Applications in Coastal Habitats},
  booktitle = {Aerial Robots -- Aerodynamics, Control and Applications},
  editor    = {Omar Dario Lopez Mejia and Jaime Alberto Escobar Gomez},
  publisher = {IntechOpen},
  year      = {2017},
  pages     = {165--184},
  doi       = {10.5772/intechopen.69598}
}

@misc{PX4,
  title        = {{PX4} Autopilot},
  howpublished = {\url{https://px4.io/}},
  year         = {2026}
}

@misc{Ardupilot,
  title        = {ArduPilot},
  howpublished = {\url{https://ardupilot.org/}},
  year         = {2026}
}

@misc{iNav,
  title        = {{INAV}: Navigation-Enabled Flight Control Software},
  howpublished = {\url{https://github.com/iNavFlight/inav}},
  year         = {2026}
}

@misc{betaflight,
  title        = {Betaflight},
  howpublished = {\url{https://www.betaflight.com/}},
  year         = {2026}
}

@article{soa_1,
  author  = {Daniele Ventura and Andrea Bonifazi and Maria Flavia Gravina and Andrea Belluscio and Giandomenico Ardizzone},
  title   = {Mapping and Classification of Ecologically Sensitive Marine Habitats Using Unmanned Aerial Vehicle ({UAV}) Imagery and Object-Based Image Analysis ({OBIA})},
  journal = {Remote Sensing},
  year    = {2018},
  volume  = {10},
  number  = {9},
  pages   = {1331},
  doi     = {10.3390/rs10091331}
}

@article{soa_2,
  author  = {J. Lindsay and J. Ross and M. L. Seto and E. Gregson and A. Moore and J. Patel and R. Bauer},
  title   = {Collaboration of Heterogeneous Marine Robots Toward Multidomain Sensing and Situational Awareness on Partially Submerged Targets},
  journal = {IEEE Journal of Oceanic Engineering},
  year    = {2022},
  volume  = {47},
  number  = {4},
  pages   = {880--894},
  doi     = {10.1109/JOE.2022.3156631}
}

@article{soa_3,
  author  = {Daniel F. Carlson and Serkan Akbulut and Jeppe Fogh Rasmussen and Christian S{\o}nderg{\aa}rd Hestbech and Marius Hjorth Andersen and Claus Melvad},
  title   = {Compact and Modular Autonomous Surface Vehicle for Water Research: The Naval Operating Research Drone Assessing Climate Change ({NORDACC})},
  journal = {HardwareX},
  year    = {2023},
  volume  = {15},
  pages   = {e00453},
  doi     = {10.1016/j.ohx.2023.e00453}
}

@article{soa_4,
  author  = {Pierre Gogendeau and Sylvain Bonhommeau and Hassen Fourati and Mohan Julien and Matteo Contini and Thomas Chevrier and Anne Elise Nieblas and Serge Bernard},
  title   = {An Autonomous Surface Vehicle for Acoustic Tracking, Bathymetric and Photogrammetric Surveys},
  journal = {Ocean Engineering},
  year    = {2025},
  volume  = {331},
  pages   = {121201},
  doi     = {10.1016/j.oceaneng.2025.121201}
}

@inproceedings{soa_6,
  author    = {Xuesu Xiao and Jan Dufek and Tim Woodbury and Robin R. Murphy},
  title     = {{UAV} Assisted {USV} Visual Navigation for Marine Mass Casualty Incident Response},
  booktitle = {2017 IEEE/RSJ International Conference on Intelligent Robots and Systems (IROS)},
  year      = {2017},
  pages     = {6105--6110},
  doi       = {10.1109/IROS.2017.8206510}
}

@article{soa_7,
  author  = {Tao Huang and Zhe Chen and Wang Gao and Zhenfeng Xue and Yong Liu},
  title   = {A {USV}-{UAV} Cooperative Trajectory Planning Algorithm with Hull Dynamic Constraints},
  journal = {Sensors},
  year    = {2023},
  volume  = {23},
  number  = {4},
  pages   = {1845},
  doi     = {10.3390/s23041845}
}

@article{soa_last,
  author  = {Dinesh Manoharan and C. Gajendran and M. K. Padmanabhan and S. Vignesh and S. Rajesh and G. D. Bhuvaneshwaran},
  title   = {Design and Development of Autonomous Amphibious Unmanned Aerial Vehicle for In Situ Water Quality Assessment and Water Sampling},
  journal = {Journal of Unmanned Vehicle Systems},
  year    = {2021},
  volume  = {9},
  number  = {3},
  pages   = {182--204},
  doi     = {10.1139/juvs-2020-0036}
}

@misc{QGC,
  title        = {QGroundControl},
  howpublished = {\url{https://qgroundcontrol.com/}},
  year         = {2026}
}

@article{allocation,
  author  = {Tor A. Johansen and Thomas P. Fuglseth and Petter T{\o}ndel and Thor I. Fossen},
  title   = {Optimal Constrained Control Allocation in Marine Surface Vessels with Rudders},
  journal = {Control Engineering Practice},
  year    = {2008},
  volume  = {16},
  number  = {4},
  pages   = {457--464},
  doi     = {10.1016/j.conengprac.2007.01.012}
}

@article{path_following,
  author  = {Claudio Paliotta and Erjen Lefeber and Kristin Y. Pettersen and Jo{\~a}o Pinto and M{\'a}rio Costa and Jo{\~a}o Tasso de Figueiredo Borges de Sousa},
  title   = {Trajectory Tracking and Path Following for Underactuated Marine Vehicles},
  journal = {IEEE Transactions on Control Systems Technology},
  year    = {2019},
  volume  = {27},
  number  = {4},
  pages   = {1423--1437},
  doi     = {10.1109/TCST.2018.2834518}
}

@inproceedings{c14,
  author    = {Brian Bingham and Carlos Ag{\"u}ero and Michael McCarrin and Joseph Klamo and Joshua Malia and Kevin Allen and Tyler Lum and Marshall Rawson and Rumman Waqar},
  title     = {Toward Maritime Robotic Simulation in {Gazebo}},
  booktitle = {OCEANS 2019 MTS/IEEE SEATTLE},
  year      = {2019},
  pages     = {1--10},
  doi       = {10.23919/OCEANS40490.2019.8962724}
}

@article{c15,
  author  = {Morten Bech Kramer and Jacob Andersen and Sarah Thomas and Flemming Buus Bendixen and Harry Bingham and Robert Read and Nikolaj Holk and Edward Ransley and Scott Brown and Yi-Hsiang Yu and Thanh Toan Tran and Josh Davidson and Csaba Horvath and Carl-Erik Janson and Kim Nielsen and Claes Eskilsson},
  title   = {Highly Accurate Experimental Heave Decay Tests with a Floating Sphere: A Public Benchmark Dataset for Model Validation of Fluid--Structure Interaction},
  journal = {Energies},
  year    = {2021},
  volume  = {14},
  number  = {2},
  pages   = {269},
  doi     = {10.3390/en14020269}
}

@misc{c10,
  author       = {Jerry Tessendorf},
  title        = {Simulating Ocean Water},
  howpublished = {SIGGRAPH Course Notes, \textit{Simulating Nature: Realistic and Interactive Techniques}},
  year         = {2001},
  note         = {SIGGRAPH 2001}
}

@article{shipdomain,
  author  = {Rafa{\l} Sz{\l}apczy{\'n}ski and Przemys{\l}aw Krata and Joanna Sz{\l}apczy{\'n}ska},
  title   = {Ship Domain Applied to Determining Distances for Collision Avoidance Manoeuvres in Give-Way Situations},
  journal = {Ocean Engineering},
  year    = {2018},
  volume  = {165},
  pages   = {43--54},
  doi     = {10.1016/j.oceaneng.2018.07.041}
}

@ARTICLE{10923704,
  author={Farinha, André and Romanello, Luca and Zufferey, Raphael and Lawson, Jenna and Armanini, Sophie Franziska and Kovac, Mirko},
  journal={IEEE Transactions on Field Robotics}, 
  title={SailMAV: Water-Surface Locomotion and Biodiversity Monitoring}, 
  year={2025},
  volume={2},
  number={},
  pages={208-229},
  doi={10.1109/TFR.2025.3547309}}

\end{document}